\documentclass{llncs}
\usepackage{graphicx}
\usepackage{mathtools}
\usepackage{prooftree}
\usepackage{url}
\usepackage{stmaryrd}
\usepackage{verbatim}
\usepackage{longtable}
\usepackage{latexsym} 
\usepackage{rotating} 
\newcommand{\typearrow}{\shortrightarrow}

\begin{document}

\title{Embedding and Automating Conditional Logics in Classical
  Higher-Order Logic}
\author{Christoph Benzm\"uller$^1$ \and Dov Gabbay$^{2}$ \and \\Valerio Genovese$^{3,4}$ \and Daniele Rispoli$^{4}$}
\institute{$^1$ Freie Universit\"at Berlin, Germany\\
$^2$ King's College London, UK\\
$^3$ University of Luxembourg, Luxembourg\\
$^4$ University of Torino, Italy}
\maketitle

\begin{abstract}
  A sound and complete embedding of conditional logics into
  classical higher-order logic is
  presented. This embedding enables the application of off-the-shelf
  higher-order automated theorem provers and model finders for
  reasoning within and about conditional logics.
\end{abstract}

\section{Introduction}
Conditional logics capture default entailment in a modal framework via
a defeasible implication operator $``\Rightarrow"$ such that $\alpha \Rightarrow \beta$ reads as, ``If $\alpha$ then, typically $\beta$". 
A peculiarity of conditional logics is that $\alpha$ is a formula and can contain other occurrences of ``$\Rightarrow$".
%

Thanks to their expressivity, conditional logics have been successfully applied in several domains like non-monotonic reasoning \cite{Boutilier94a}, belief revision \cite{CroccoL92} and security 
\cite{GenoveseGGP10}.

Despite their wide range of potential applications, the formalization
of a proper proof theory for conditional logics has been tackled only
recently and only for a limited set of axiomatizations
\cite{Pozzato10,Pattinson10}. Moreover, there is no uniform framework
to specify and reason about such formalisms.


Following the work of \cite{J21}, a semantic embedding of conditional
logic in classical higher-order logic HOL (Church's type theory) is
presented. This embedding exploits the natural correspondence
between selection function semantics for conditional logics \cite{Sta68}
and HOL.
In fact, selection function semantics can be seen as an higher-order extension of well-known
Kripke semantics for modal logic and cannot be naturally embedded into first-order logic. 

The contributions of the paper are threefold.
First, we prove that the presented embedding is \emph{sound} and \emph{complete} w.r.t.
selection function semantics.
Second, we show how to apply off-the-shelf higher-order theorem provers
and model finders for reasoning \emph{within} and \emph{about} conditional logics.
Third, we investigate the practical value of such embedding through several experiments
with different higher-order reasoning systems (HOL-RSs). As part of these experiments,
several correspondence results between prominent conditional logic axioms and related
semantic conditions have been automatically verified.




\section{Conditional Logics}\label{sec:conditional_logics}

In order to make the paper self contained, we briefly resume syntax
and semantics of conditional logics. For a deeper treatment we refer
to \cite{Pozzato10}.
%
%

\begin{definition} The \emph{formulas} of conditional logic are given by
$$\varphi ::= p \mid \neg \varphi \mid 
\varphi \vee \varphi \mid \varphi \Rightarrow \varphi$$
where $p$ ranges over a set of Boolean variables and $\Rightarrow$ is a binary modal operator.
\end{definition}
%
From the selected set of primitive connectives, other logical connectives can be
introduced as abbreviations: e.g., $\varphi\wedge\psi$ and
$\varphi \rightarrow \psi$ (material implication) 
abbreviate  $\neg (\neg \varphi \vee \neg \psi)$ and $\neg
\varphi \vee \psi$, etc.  Syntactically, conditional logics can be seen
as a generalization of multimodal logic where the index of modality
$\Rightarrow$ is a formula of the same language. For instance, in $(A \Rightarrow B) \Rightarrow C$
the subformula $A \Rightarrow B$ is the index of the second occurrence
of $\Rightarrow$. 

Regarding semantics, many different formalizations have been
proposed (see \cite{Nut80}),
here we focus on the \emph{selection function semantics}
\cite{Chellas80}, which is based on possible world structures and 
has been successfully used in \cite{Olivetti07} to develop proof methods for some conditional logics. 

\begin{definition}\label{def:CKmodels}
A \emph{model} of conditional logics  is a tuple $\mathcal{M} = \langle S, f, h\rangle$
where,
\begin{itemize}
\item $S$ is a non empty set of items called states;
\item $f: S \times 2^{S} \rightarrow 2^S$ is the selection function;
\item $h$ is an assignment which, for each Boolean variable $p$, assigns the subset of states $h(p)$ where $p$ holds. 
\end{itemize}
\end{definition}
Intuitively the selection function $f$ selects, for a world $w$ and a formula $\varphi$, the set of worlds $f(w,A)$ which are 
``most-similar to $w$" or ``closer to $w$" given the information $\varphi$.

\begin{definition}[Semantic Interpretation]
  An interpretation for a conditional logic is a pair $\mathcal{M},s$
  where $\mathcal{M}$ is a model and $s$ is a state in
  $\mathcal{M}$. The satisfaction relation $\models$ holds between
  interpretations and formulae of the logic, and it is defined
  recursively as follows:
\begin{itemize}
\item $\mathcal{M},s \models p$ iff $s \in h(p)$
\item $\mathcal{M},s \models \neg \varphi$ iff  not $\mathcal{M},s \models \varphi$
\item $\mathcal{M},s \models \varphi \vee \psi$ iff $\mathcal{M},s \models \varphi$ or $\mathcal{M},s \models \psi$
\item $\mathcal{M},s \models \varphi \Rightarrow \psi$ iff $\mathcal{M},t \models \psi$  for all $t \in f(s,[\varphi])$  where,
$[\varphi] = \{w \mid \mathcal{M},w \models \varphi\}$.
\end{itemize}
\end{definition}
As usual, a conditional formula $\varphi$ is \emph{valid in a model} $\mathcal{M}=\langle S,f,h \rangle$, denoted with $\mathcal{M} \models \varphi$, iff for all $s \in S$ holds $\mathcal{M},s \models \varphi$. A formula $\varphi$ is  \emph{valid}, denoted $\models \varphi$, iff it is valid in every model.

Notice that $f$ is defined to take $[\varphi]$ (called the \emph{proof set} of $\varphi$ w.r.t. a given model $\mathcal{M}$) instead
of $\varphi$. This approach has the consequence of forcing the so-called \emph{normality} property: given a model $\mathcal{M}$, 
if $\varphi$ and $\varphi'$ are equivalent (i.e., they are satisfied in the same set of states), then they index the same formulas w.r.t. to the $\Rightarrow$ modality.

The axiomatic counterpart of the normality condition is given by the
rule (RCEA)
\begin{center}
\begin{prooftree}
  \varphi \leftrightarrow \varphi' \justifies (\varphi \Rightarrow
  \psi) \leftrightarrow (\varphi' \Rightarrow \psi) \using (RCEA)
\end{prooftree}
\end{center}
Moreover, it can be easily shown that the above semantics forces also
the following rules to hold:
\begin{center}
\begin{prooftree}
  (\varphi_1 \wedge \ldots \wedge \varphi_n) \leftrightarrow \psi
  \justifies (\varphi_0 \Rightarrow \varphi_1 \wedge \ldots \wedge
  \varphi_0 \Rightarrow \varphi_n) \rightarrow (\varphi_0 \Rightarrow
  \psi) \using (RCK)
\end{prooftree}
\end{center}
\begin{center}
\begin{prooftree}
\varphi \leftrightarrow \varphi'
\justifies (\psi \Rightarrow \varphi) \leftrightarrow (\psi \Rightarrow \varphi')  \using (RCEC)
\end{prooftree}
\end{center}

We refer to $CK$ \cite{Chellas80} as the minimal conditional logic
closed under rules RCEA, RCEC and RCK.  In what follows, only
conditional logics extending CK are considered.
\section{Classical Higher-Order Logic}\label {sec:hol}
HOL is a logic based on simply typed $\lambda$-calculus
\cite{Church40,AndrewsSEP}.  The set $\mathcal{T}$ of simple types in
HOL is usually freely generated from a set of basic types $\{o,i\}$
(where $o$ denotes the type of Booleans) using the function type
constructor $\typearrow$.

\begin{definition}
The \emph{terms} of HOL are defined by ($\alpha,\beta,o \in \mathcal{T}$)
\[
\begin{array}{ll}
s,t ::= & p_{\alpha} \mid X_{\alpha} \mid (\lambda X_{\alpha}.s_{\beta})_{\alpha \typearrow \beta} \mid 
(s_{\alpha \typearrow \beta})_{\beta} \mid (\neg_{o \typearrow o}\;s_{o})_o \mid \\ 
& (s_o \vee_{o\typearrow o \typearrow o} t_o)_o 
\mid (\Pi_{(\alpha \typearrow o)\typearrow o}\; s_{\alpha \typearrow o})_o
\end{array}
\]
$p_{\alpha}$ denotes typed constants and $X_{\alpha}$ typed variables
(distinct from $p_\alpha$). 
\end{definition}
Complex typed terms are constructed via abstraction and
application. The primitive logical connectives are $\neg_{o \typearrow
  o},\vee_{o \typearrow o \typearrow o}$ and $\Pi_{(\alpha \typearrow
  o) \typearrow o}$ (for each type $\alpha$). From these, other
logical connectives can be introduced as abbreviations: e.g., $\wedge$
and $\rightarrow$ abbreviate the terms $\lambda A. \lambda B. \neg
(\neg A \vee \neg B)$ and $\lambda A. \lambda B. \neg A \vee B$,
etc. HOL terms of type $o$ are called formulas.
Binder notation $\forall X_{\alpha}.s_o$ is used as an abbreviation
for ($\Pi_{(\alpha \typearrow o)\typearrow o}\;(\lambda
X_{\alpha}. s_{o})$).  Substitution of a term $A_{\alpha}$ for a
variable $X_{\alpha}$ in a term $B_{\beta}$ is denoted by $[A/X]B$, where it is
assumed that the bound variables of $B$ avoid variable capture. Well known operations and relations on HOL terms 
include $\beta\eta$-normalization and $\beta\eta$-equality, denoted by  $s =_{\beta\eta} t$. 

The following definition of HOL semantics 
closely follows the standard literature \cite{Andrews1972b,AndrewsSEP}.

\begin{definition}
\label{stt_frame}
A \emph{frame} is a collection $\{D_{\alpha}\}_{\alpha \in \mathcal{T}}$ of nonempty sets called \emph{domains} such that $D_{o} = \{T,F\}$ where $T$ represents truth and $F$ falsehood, $D_{i}\not=\emptyset$ is chosen arbitrary, and $D_{\alpha \typearrow \beta}$ are collections of functions mapping $D_{\alpha}$ into $D_{\beta}$.
\end{definition}
\begin{definition}\label{stt_interpretation}
  An \emph{interpretation} is a tuple $\langle \{D_{\alpha}\}_{\alpha \in
    \mathcal{T}}, I \rangle$ where $\{D_{\alpha}\}_{\alpha \in
    \mathcal{T}}$ is a frame and where function $I$ maps each typed
  constant $c_{\alpha}$ to an appropriate element of $D_{\alpha}$,
  which is called the \emph{denotation} of $c_{\alpha}$. The
  denotations of $\neg,\vee$ and $\Pi_{(\alpha \typearrow o)\typearrow
    o}$ are always chosen as usual. A variable assignment $\phi$ maps
  variables $X_{\alpha}$ to elements in $D_{\alpha}$. An
  interpretation is a \emph{Henkin model (general model)} if and only
  if there is a binary valuation function $\mathcal{V}$ such that
  $\mathcal{V}(\phi,s_{\alpha}) \in D_{\alpha}$ for each variable
  assignment $\phi$ and term $s_{\alpha}$, and the following
  conditions are satisfied for all $\phi$, variables $X_\alpha$,
  constants $p_\alpha$, and terms $l_{\alpha \typearrow \beta},
  r_\alpha, s_\beta$ (for $\alpha,\beta\in\mathcal{T}$):

\begin{itemize}
\item $\mathcal{V}(\phi,X_{\alpha}) = \phi(X_\alpha)$
\item $\mathcal{V}(\phi,p_{\alpha}) = I(p_{\alpha})$
\item $\mathcal{V}(\phi,(l_{\alpha \typearrow \beta}\; r_{\alpha})) = (\mathcal{V}(\phi,l_{\alpha \typearrow \beta}))(\mathcal{V}(\phi,r_{\alpha})) $
\item $\mathcal{V}(\phi,\lambda X_{\alpha}.s_{\beta})$ represents the function from $D_{\alpha}$ into $D_{\beta}$ whose value for each argument
$z \in D_{\alpha}$ is $\mathcal{V}(\phi[z/X_{\alpha}], s_\beta)$, where $\phi[z/X_\alpha]$ is that variable assignment such that $\phi[z/X_\alpha](X_\alpha) = z$
and $\phi[z/X_\alpha]Y_\beta = \phi Y_\beta$ when $Y_\beta \not=X_\alpha$.
\end{itemize}
\end{definition}

If an interpretation $\mathcal{H} = \langle \{D_\alpha\}_{\alpha \in
  \mathcal{T}}, I\rangle$ is an \emph{Henkin model} the function
$\mathcal{V}$ is uniquely determined and
$\mathcal{V}(\phi,s_\alpha)\in D_\alpha$ is called the denotation of
$s_\alpha$. $\mathcal{H}$ is called a \emph{standard model} if and
only if for all $\alpha$ and $\beta$, $D_{\alpha \typearrow \beta}$ is
the set of all functions from $D_{\alpha}$ into $D_{\beta}$. It is
easy to verify that each standard model is also a Henkin model. A
formula $A$ of HOL is \emph{valid} in a Henkin model $\mathcal{H}$ if
and only if $\mathcal{V}(\phi,A) = T$ for all variable assignments
$\phi$. In this case we write $\mathcal{H}\models A$.  $A$ is (Henkin) valid,
denoted as $\models A$, if and only if $\mathcal{H}\models A$ for all 
Henkin models 
$\mathcal{H}$.

\begin{proposition}\label{prop:trivial}
  Let $\mathcal{V}$ be the valuation function of Henkin model
  $\mathcal{H}$. The following properties hold for all 
  assignments $\phi$, terms $s_o,t_o,l_\alpha,r_\alpha$, and variables
  $X_\alpha,V_\alpha$ (for $\alpha\in\mathcal{T}$):
\begin{itemize}
\item $\mathcal{V}(\phi,(\neg s_o))=T$ iff $\mathcal{V}(\phi,s_o)=F$
\item $\mathcal{V}(\phi,(s_o \vee t_o))=T$ iff $\mathcal{V}(\phi,s_o)=T$ or $\mathcal{V}(\phi,s_o)=T$
\item $\mathcal{V}(\phi,(s_o \wedge t_o))=T$ iff $\mathcal{V}(\phi,s_o)=T$ and $\mathcal{V}(\phi,s_o)=T$
\item $\mathcal{V}(\phi,(s_o \rightarrow t_o))=T$ iff $\mathcal{V}(\phi,s_o)=F$ or $\mathcal{V}(\phi,s_o)=T$
\item $\mathcal{V}(\phi,(\forall X_\alpha. s_o))=
  \mathcal{V}(\phi,(\Pi_{(\alpha\typearrow o) \typearrow o}\; (\lambda
  X_\alpha. s_o)))=T$ iff  for all $v\in D_\alpha$ holds $\mathcal{V}(\phi[v/V_\alpha],((\lambda
  X_\alpha. s_o)\;V))=T$
\item if $l_\alpha =_{\beta\eta} r_\alpha$ then $\mathcal{V}(\phi,l_\alpha)=\mathcal{V}(\phi,r_\alpha)$
\end{itemize}
\end{proposition}

\section{Embedding Conditional Logics in HOL}\label{sec:embedding}
Conditional logic formulas
are identified with certain HOL terms (predicates) of type $i
\typearrow o$. They can be applied to terms of type $i$, which are
assumed to denote possible states.

\begin{definition}\label{def:embedding}
  The mapping $\lfloor \cdot \rfloor$ translates
  formulas $\varphi$ of conditional logic $CK$ into HOL terms $\lfloor
  \varphi \rfloor$ of type $i \typearrow o$. The primitives of
  conditional logic are mapped as follows:
\[
\begin{array}{ll}
\lfloor p\rfloor &= p_{i \typearrow o}  \qquad \qquad (p_{i \typearrow o}\; \text{is a HOL constant symbol)}\\
\lfloor\neg\rfloor &= \lambda A_{i \typearrow o}.\lambda X_{i}. \neg(A\; X)\\
\lfloor\vee\rfloor &= \lambda A_{i \typearrow o}.\lambda B_{i \typearrow o}.\lambda X_i. (A\;X) \vee (B\;X)\\
\lfloor\Rightarrow\rfloor &= \lambda A_{i \typearrow o}.\lambda B_{i \typearrow o}.\lambda X_{i}.\forall W_i.(f\; X\; A\; W) \rightarrow (B\; W)\\
\end{array}
\]
The constant symbol $f$ in the mapping of $\Rightarrow$ is of type ${i
  \typearrow (i \typearrow o) \typearrow (i \typearrow o)}$. It
realizes the selection function, i.e., its interpretation is
chosen appropriately (cf. below).

Compound formulas are recursively mapped as follows:
\[
\begin{array}{ll}
\lfloor \neg \varphi \rfloor &=  (\lfloor \neg \rfloor \lfloor \varphi \rfloor) \\
\lfloor \varphi\vee \psi \rfloor &=  (\lfloor \vee \rfloor \lfloor\varphi \rfloor \lfloor \psi \rfloor)\\
\lfloor \varphi\wedge \psi \rfloor &=  (\lfloor \wedge\rfloor \lfloor\varphi \rfloor \lfloor \psi \rfloor)\\
\lfloor \varphi\Rightarrow \psi \rfloor &=  (\lfloor \Rightarrow \rfloor  \lfloor\varphi \rfloor \lfloor \psi \rfloor)
\end{array}
\]

Analyzing the validity of a translated formula $\lfloor \varphi
\rfloor$ for a state represented by term $t_i$ corresponds
to evaluating the application $(\lfloor \varphi \rfloor\;t_i)$. In
line with \cite{J21}, we can easily encode the notion of validity as
follows
$$ vld := \lambda A_{i \typearrow o}. \forall S_{i}.(A\; S)$$ 
With this definition, validity of a conditional formula $\varphi$ in
CK corresponds to the validity of the corresponding formula $
(\text{vld}\; \lfloor \varphi \rfloor)$ in HOL, and vice versa.
\end{definition}

We illustrate the approach with formula $p \Rightarrow p$
where $p$ is a Boolean variable.  This formula
corresponds to the HOL term $(\text{vld}\; \lfloor p \Rightarrow
p\rfloor)$ which expands into (type information is omitted) $ (\lambda
A. \forall S.(A\; S)) ((\lambda A.\lambda B.\lambda X.\forall W.(f\;
X\; A\; W) \rightarrow (B\; W)) \; p\; p) $ and $\beta\eta$-normalizes to
$\forall S. \forall W.(f\; S\; p\; W) \rightarrow (p\; W))$.  It is
easy to verify that this HOL formula is countersatisfiable, which is
the expected result in CK.

To prove the soundness and completeness of the embedding, a
mapping from selection function models into Henkin models is employed.

\begin{definition}\label{embedding}
  Given a selection function model $\mathcal{M} = \langle S, f,\models\rangle$. The Henkin model 
  $\mathcal{H}^{\mathcal{M}} = \langle \{D_\alpha\}_{\alpha \in
    \mathcal{T}}, I \rangle$ for $\mathcal{M}$ is defined as follows:
  $D_i$ is chosen as the set of states $S$, and
  for all combinations of $\alpha$ and $\beta$, $D_{\alpha
    \typearrow \beta}$ is chosen as the set of all functions from $D_\alpha$ to
  $D_\beta$.\footnote{This choice in particular means that $D_{i \typearrow
    o}$ is the set of all possible predicates $q$ over $S$; these
  predicates can also be viewed a sets $\{x\in S\mid q(x) =
  T\}$. Note, that modulo this technicality, $D_{i \typearrow o}$ is identical to '$2^S$' in
  Def.~\ref{def:CKmodels}.}  Let $p^1,\ldots,p^m$ for $m \geq 1$
  be the Boolean variables of the conditional logic and let $\lfloor p^{j} \rfloor = p^{j}_{i
    \typearrow o}$ for $i=1,\ldots,m$.  We define $I$ as follows:
\begin{itemize}
\item For $1 \leq i \leq m$, choose $I(p^{j}_{i \typearrow o}) \in
  D_{i \typearrow o}$ so that $(I(p^{j}_{i \typearrow o}))(w) = T$
  for all $w \in D_i$ with $\mathcal{M},w \models p^j$, and
  $(I(p^{j}_{i \typearrow o}))(w) = F$ otherwise.
\item Choose $I(f_{i \typearrow (i \typearrow o) \typearrow (i
    \typearrow o)}) \in D_{i \typearrow (i \typearrow o) \typearrow (i
    \typearrow o)}$ so that for all $s,t\in D_i$ and $q\in D_{i
    \typearrow o}$ holds $(I(f_{i \typearrow (i \typearrow o)
    \typearrow (i \typearrow o)}))(s,q,t) = T$ if $t \in f(s,\{x\in S\mid
  q(x) = T\})$ in $\mathcal{M}$, and $(I(f_{i \typearrow (i
    \typearrow o) \typearrow (i \typearrow o)}))(s,q,t) = F$
  otherwise.
\item 
  For all other constants $s_\alpha$, choose $I(s_\alpha)$ arbitrarily.\footnote{In fact, we may safely assume that there are
  no other typed constant symbols given, except for the symbol $f_{i
    \typearrow (i \typearrow o) \typearrow (i \typearrow o)}$, the
  symbols $p^{j}_{i \typearrow o}$, and the logical connectives.}
\end{itemize}
\end{definition}
It is easy to verify that $\mathcal{H}^{\mathcal{M}}$ is a Henkin
model.\footnote{In $\mathcal{H}^{\mathcal{M}}$ we have merely fixed
  $D_i$ and the interpretation of the constant symbols $p^j_{i
    \typearrow o}$ and $f_{i \typearrow (i \typearrow o) \typearrow (i
    \typearrow o)}$.  These constraints are obviously not in conflict
  with any of the requirements in Defs.~\ref{stt_frame} and
  \ref{stt_interpretation}. The existence of a valuation function
  $\mathcal{V}$ for an HOL interpretation crucially depends on how
  sparse the function spaces have been chosen in frame
  $\{D_\alpha\}_{\alpha \in \mathcal{T}}$. \cite{Andrews1972b}
  discusses criteria that are sufficient to ensure the existence of a
  valuation function; they require that certain $\lambda$-abstractions
  have denotations in frame $\{D_\alpha\}_{\alpha \in
    \mathcal{T}}$. The function spaces are full, so this is trivially
  the case.} It is even a standard model, since the function spaces
are full.

\begin{lemma}\label{lemma:1}
  Let $\mathcal{H}^{\mathcal{M}}$ be a Henkin model for a selection
  function model $\mathcal{M}$. For all conditional logic formulas
  $\varphi$, states $s$, and variable assignments $\phi$
  it holds: $$\mathcal{M},s \models \varphi \text{\quad iff \quad}
  \mathcal{V}(\phi[s/S_i], (\lfloor \varphi\rfloor\; S))=T$$
\begin{proof}
  The proof is by induction on the structure of $\varphi$.

  $\varphi = p^j$.  By definition of $\lfloor \cdot \rfloor$,
  $\mathcal{V}$, and $\mathcal{H}^{\mathcal{M}}$ holds
  $\mathcal{V}(\phi[s/S_{i}], (\lfloor p^j\rfloor\; S))=
  \mathcal{V}(\phi[s/S_i], (p^j_{i \typearrow o}\; S)) = (I
  (p^j_{i\typearrow o}))(s) = T$ iff $\mathcal{M}, s \models p^j$.

    For $\varphi = (\neg r)$ the argument is similar to $\varphi = (p \vee r)$ below.

  $\varphi = (p \vee r)$. $\mathcal{M}, s \models (p \vee t)$ iff
  $\mathcal{M}, s \models p$ or $\mathcal{M}, s \models t$. By
  induction $\mathcal{V}(\phi[s/S_i], (\lfloor p\rfloor\;S)) = T$ or
  $\mathcal{V}(\phi[s/S_i], (\lfloor r\rfloor\;S)) = T$. By
  Prop.~\ref{prop:trivial}, definition of $\lfloor\cdot\rfloor$ and
  since $((\lfloor p \vee r \rfloor)\;S)=_{\beta\eta} ((\lfloor
  p\rfloor\;S) \vee (\lfloor r \rfloor\; S))$ it holds
  $\mathcal{V}(\phi[s/S_i],((\lfloor p \vee r\rfloor)\;S)) =
  \mathcal{V}(\phi[s/S_i],((\lfloor p\rfloor\;S) \vee (\lfloor r
  \rfloor\; S))) = T$.

  $\varphi = (p \Rightarrow r)$.  $\mathcal{M}, s \models p
  \Rightarrow r$ iff, for all $t \in f(s,[p])$
  holds $\mathcal{M}, t \models r$. This is equivalent to, for all $t$ holds
  (i) $t\not\in f(s,[p])$ or (ii) $\mathcal{M}, t
  \models r$.  By induction applied to (ii) with $\phi'=\phi[s/S_i]$
  holds $\mathcal{V}(\phi'[t/T_i],(\lfloor r\rfloor\; T)) = T$. Furthermore, again by induction, 
  for all $t$ and $\phi''$ with
  $\phi''=\phi'[t/T_i]=\phi[s/S_i][t/T_i]$ we have
  $\mathcal{M},u\models p$ iff
  $\mathcal{V}(\phi''[u/U_i],(\lfloor p\rfloor\; U_i)) = T$. Hence, by
  construction of $\mathcal{H}^\mathcal{M}$, (i) is equivalent to
  $(I(f))(s,[p],t) = (I(f))(s,\{u\mid \mathcal{M},u\models
  p\},t) = (I(f))(s,\{u\mid \mathcal{V}(\phi''[u/U_i],(\lfloor
  p\rfloor\; U_i)) = T\},t) = F $. By definition of $\mathcal{V}$,
  and since $s=\mathcal{V}(\phi'',S_i)$ and
  $t=\mathcal{V}(\phi'',T_i)$ it holds
  $\mathcal{V}(\phi'',(f\;S_i\;\lfloor p\rfloor\;T_i)) = F$. 
  By combining these results and by Prop.~\ref{prop:trivial} we get that for all $t$
  $\mathcal{V}(\phi'[t/T_i],((f\;S_i\;\lfloor p\rfloor\;T_i)
  \rightarrow (\lfloor r\rfloor\;T_i))) =
  \mathcal{V}(\phi'[t/T_i],(\lambda W. ((f\;S_i\;\lfloor
  p\rfloor\;W_i) \rightarrow (\lfloor r\rfloor\;W_i))\;T_i))= T$, which by Prop.~\ref{prop:trivial} is equivalent \sloppy
  to $\mathcal{V}(\phi', (\Pi_{(i \typearrow o)\typearrow o}\;
  (\lambda W_i. ((f\;S_i\;\lfloor p\rfloor\;W_i) \rightarrow (\lfloor
  r\rfloor\;W_i))))) = \mathcal{V}(\phi',(\forall
  W_i.((f\;S_i\;\lfloor p\rfloor\;W_i) \rightarrow (\lfloor
  r\rfloor\;W_i))) =T$. By  Prop.~\ref{prop:trivial}, definition of $\lfloor\cdot\rfloor$ and
  since  $(\lfloor p \Rightarrow r
  \rfloor\;S)=_{\beta\eta} (\forall W_i.((f\;S_i\;\lfloor
  p\rfloor\;W_i) \rightarrow (\lfloor r\rfloor\;W_i)))$ we finally have
  $\mathcal{V}(\phi[s/S_i],(\lfloor p \Rightarrow r \rfloor\;S)) = T$.
\end{proof}
\end{lemma}

\begin{theorem}[Soundness and Completeness]\label{sound_compl}
  $$\models (\text{vld}\; \lfloor \varphi \rfloor) \text{ in HOL \quad  if and
  only if \quad} \models \varphi \text{ in CK}$$
\begin{proof} (Soundness) The proof is by contraposition. Assume
  $\not\models \varphi$ in CK, that is, there is a selection function
  model $\mathcal{M}=\langle S,f,h \rangle$ and a state $s\in S$, such
  that $\mathcal{M},s\not\models \varphi$. By Lemma \ref{lemma:1} we
  have that $\mathcal{V}(\phi[s/S_i],(\lfloor \varphi \rfloor\;S)) =
  F$ for a variable assignment $\phi$ in Henkin model
  $\mathcal{H}^\mathcal{M}= \langle \{D_\alpha\}_{\alpha \in
    \mathcal{T}}, I \rangle$ for $\mathcal{M}$. Thus, by
  Prop.~\ref{prop:trivial}, definition of $\text{vld}$ and since
  $(\forall S_i. \lfloor \varphi \rfloor\;S) =_{\beta\eta}
  (\text{vld}\;\lfloor \varphi \rfloor)$ we know that
  $\mathcal{V}(\phi,(\forall S_i. \lfloor \varphi \rfloor\;S)) =
  \mathcal{V}(\phi,(\text{vld}\;\lfloor \varphi \rfloor)) =
  F$. Hence, $\mathcal{H}^\mathcal{M}\not\models
  (\text{vld}\;\lfloor \varphi \rfloor)$, and thus $\not\models
  (\text{vld}\;\lfloor \varphi \rfloor)$ in HOL.

  (Completeness) The proof is again by contraposition. Assume $\not\models (\text{vld}\;
  \lfloor \varphi \rfloor)$ in HOL, that is, there is a Henkin model
  $\mathcal{H}=\langle \{D_\alpha\}_{\alpha \in \mathcal{T}}, I
  \rangle$ and a variable assignment $\phi$ with
  $\mathcal{V}(\phi,(\text{vld}\; \lfloor \varphi \rfloor)) = F$. From
  $(\text{vld}\;\lfloor \varphi \rfloor) =_{\beta\eta} (\forall
  S_i. \lfloor \varphi \rfloor\;S)$ and 
  Prop.~\ref{prop:trivial} we get
  $\mathcal{V}(\phi,(\forall S_i. \lfloor \varphi \rfloor\;S)) = F$, 
  and hence, by definition of $vld$,  $\mathcal{V}(\phi[s/S_i],\lfloor \varphi \rfloor\;S) = F$
  for some $s\in D_i$. Without loss of generality we can assume that
  Henkin Model $\mathcal{H}$ is in fact a Henkin model
  $\mathcal{H}^\mathcal{M}$ for a corresponding selection function
  model $\mathcal{M}$.  By Lemma \ref{lemma:1} we thus
  know that $\mathcal{M},s \not \models \varphi$, and hence
  $\not\models \varphi$ in CK.
\end{proof}
\end{theorem} 

Theorem \ref{sound_compl} does not trivially follow from previous literature on embedding modal logics into HOL because of the complexity of the selection function. In fact, standard modalities are usually evaluated over a so-called accessibility relationship of type $R(i,w)$, where $i$ is an index and $w$ is a world. Conditional modalities are instead evaluated over selection functions of type $f(w,A)$ where $w$ is a world, $A$ is a \emph{set of worlds} and $f$ is a \emph{function} which returns a set of worlds.

\section{Experiments: Analyzing the Literature}\label{sec:experiments}
The presented semantic embedding of conditional logics into HOL is of
practical relevance. It supports the application of standard HOL-RSs
to problems encoded \emph{within} conditional logics and also to problems
\emph{about} conditional logics. 
Examples of the latter kind include correspondence claims between
certain axioms and related conditions on the semantic structures
(e.g., the conditional logic axiom $\varphi \Rightarrow \varphi$ corresponds to
the semantic condition that $f(w,[\varphi])\subseteq
[\varphi]$). 

This section reports on experiments in which
such kind of questions have been studied with
HOL-RSs. 
The HOL-RSs employed in the case study
are:
\begin{description}
\item[\bf LEO-II (version v1.2.6).] A higher-order automated theorem prover based on extensional resolution~\cite{C26}. LEO-II\footnote{\scriptsize \url{http://www.leoprover.org}}
cooperates with the first-order theorem prover E. 
\item[\bf TPS (version 3.080227G1d).] A fully automated version of the
  higher-order theorem proving environment {TPS}\footnote{\scriptsize \url{http://gtps.math.cmu.edu/tps.html}}~\cite{AB+96}. 
  Proof search in TPS is controlled by modes (sets of flag
  settings), and the automated TPS version employed here applies strategy scheduling
  over these modes.  
\item[\bf Satallax (version 1.4).]  A higher-order automated theorem
  prover based on a complete ground tableau calculus for HOL with a
  choice operator~\cite{BackesBrown2010b}. 
  Satallax\footnote{\scriptsize \url{http://www.ps.uni-saarland.de/~cebrown/satallax/}}, which cooperates with SAT solver MiniSat, 
  has additional model finding capabilities.
\item[\bf IsabelleP (version 2009-2).] The proof
   assistant Isabelle/HOL\footnote{\scriptsize \url{http://isabelle.in.tum.de}}~\cite{NPW02} 
is normally used interactively.
   IsabelleP is an automated version of Isabelle/HOL, in
   which several tactics are subsequently applied.
\item[\bf Refute and Nitpick (versions 2009-2).]  Isabelle/HOL's ability to find \sloppy (counter-)models
  using the \emph{refute}~\cite{weber-2008} 
and
  \emph{nitpick}\footnote{\scriptsize \url{http://www4.in.tum.de/~blanchet/nitpick.html}}~\cite{DBLP:conf/itp/BlanchetteN10} 
commands has been  integrated into automatic systems. 
\end{description}

The reasoning systems described above are available online via the
System\-On\-TPTP tool \cite{DBLP:journals/jar/Sutcliffe09}. They
support the new TPTP THF infrastructure for HOL \cite{J22} and they
accept problems formalized in the THF representation language.

The problems studied in the experiments are:
\begin{problem} \label{Problem1} Is the presented embedding
  consistent? In order to study this question, the
  formalization
of the embedding
  has  been passed to the HOL-RSs.
\end{problem}

\begin{problem}\label{Problem2} Are the rules RCEA, RCK, and RCEC implied in the embedding? They obviously should, since CK is defined as the minimal conditional
  logic closed under these rules. The problems passed to the HOL-RSs
  are (types are omitted):
RCEA:  $\forall P,Q,R. (\text{vld}\;\lfloor P \leftrightarrow Q \rfloor) \rightarrow (\text{vld}\;\lfloor (P \Rightarrow R) \leftrightarrow (Q \Rightarrow R) \rfloor)$, 
RCK: $\forall P0,P1,P2,Q. (\text{vld}\;\lfloor (P1 \wedge P2) \leftrightarrow Q \rfloor) \rightarrow (\text{vld}\;\lfloor ( (P0 \Rightarrow P1) \wedge (P0 \Rightarrow P2)) \rightarrow (P0 \Rightarrow Q) \rfloor)$, RCEC: $\forall P,Q,R. (\text{vld}\;\lfloor P \leftrightarrow Q \rfloor) \rightarrow (\text{vld}\;\lfloor (R \Rightarrow P) \leftrightarrow (R \Rightarrow Q) \rfloor)$ \label{P2_3} 
\end{problem}

\begin{figure*}[t]\centering
\begin{tabular}{| c | c | c |}
\hline
\textbf{ID} & Axiom & $A \Rightarrow A$ \\
            & Condition & $f(w,[A]) \subseteq [A]$ \\ \hline
\textbf{MP} & Axiom & $(A \Rightarrow B) \rightarrow (A \rightarrow B)$ \\
            &  Condition & $w \in [A] \rightarrow w \in f(w,[A])$ \\ \hline
\textbf{CS} & Axiom & $(A \wedge B) \rightarrow (A \Rightarrow B)$ \\
            &  Condition & $w \in [A] \rightarrow f(w,[A]) \subseteq \{w\}$ \\ \hline 
\textbf{CEM} & Axiom & $(A \Rightarrow B) \vee (A \Rightarrow \neg B)$ \\
             &  Condition &  $|f(w,[A])| \leq 1$ \\  \hline
\textbf{AC} & Axiom & $(A \Rightarrow B) \wedge (A \Rightarrow C) \rightarrow (A \wedge C \Rightarrow B)$ \\
            &  Condition &  $f(w,[A]) \subseteq [B] \rightarrow f(w,[A \wedge B]) \subseteq f(w,[A])$ \\ \hline
\textbf{RT} & Axiom & $(A \wedge B \Rightarrow C) \rightarrow ((A \Rightarrow B) \rightarrow (A \Rightarrow C))$ \\
            &  Condition & $f(w,[A]) \subseteq [B] \rightarrow f(w,[A])
\subseteq f(w,[A \wedge B])$ \\ \hline
\textbf{CV} & Axiom & $(A \Rightarrow B) \wedge \neg(A \Rightarrow \neg C) \rightarrow (A \wedge C \Rightarrow B)$ \\
             &  Condition & $(f(w,[A]) \subseteq [B] \mbox{ and } f(w,[A]) \cap [C] \not= \emptyset)\rightarrow f(w,[A \wedge C]) \subseteq [B]$ \\ \hline
\textbf{CA} & Axiom & $(A \Rightarrow B) \wedge (C \Rightarrow B) \rightarrow (A \vee C \Rightarrow B)$ \\
            &  Condition & $f(w,[A \vee B]) \subseteq f(w,[A]) \cup f(w,[B])$ \\ \hline
\end{tabular}
\caption{Conditional logic axioms and semantic conditions \label{fig:correspondence}}
\end{figure*}

\begin{problem}\label{Problem3} Do the correspondence results between conditional
  logic axioms and semantic conditions as presented in
  Figure~\ref{fig:correspondence} (copied from \cite{Pozzato10}) indeed hold?
\begin{description}
\item[ID] $(\forall A.  \text{vld}\;\lfloor A \Rightarrow A \rfloor) \leftrightarrow 
  (\forall A, W. (f\;W\;A) \subseteq A)$ 
\item[MP] $(\forall A,B.  \text{vld}\;\lfloor (A \Rightarrow B)
  \rightarrow (A \rightarrow B) \rfloor) \leftrightarrow (\forall A,W. (A\;W)
  \rightarrow ((f\;W\;A)\;W))$  
\item[CS] $(\forall A,B. \text{vld}\;\lfloor (A \wedge B) \rightarrow
  (A \Rightarrow B) \rfloor) \leftrightarrow
  (\forall A,W. (A\;W) \rightarrow (f\;W\;A) \subseteq (\lambda X_i. X = W))$ 
\item[CEM] \sloppy $(\forall A,B. \text{vld}\;\lfloor (A \Rightarrow B) \vee
  (A \Rightarrow \neg B) \rfloor) \leftrightarrow
  (\forall A,W. (f\;W\;A)$\ $= \emptyset \vee \exists V. (f\;W\;A) = (\lambda X . X = V))$ 
\item[\ldots] The formalizations of AC, RT, CV, and CA are analogous, and the HOL encodings of $\subseteq$, $\emptyset$, $\cap$, and $\cup$ are
straightforward.

\end{description}
In the experiments, each equivalence statement has actually been split
in its two implication directions.
\end{problem}

\begin{problem}\label{Problem4} 
  A subtle point, concerning correspondence theory for conditional logics,
  is the interpretation of the scopes of the implicit universal quantifiers in
  the correspondence statements in Figure~\ref{fig:correspondence}.
  For example, for ID and MP we
  might read (ID') $\forall A.  ((\text{vld}\;\lfloor A \Rightarrow A
  \rfloor) \leftrightarrow W. ((f\;W\;A) \subseteq A))$ and (MP')
  $\forall A,B. (\text{vld}\;\lfloor (A \Rightarrow B) \rightarrow (A
  \rightarrow B) \rfloor \leftrightarrow (\forall W. (A\;W)
  \rightarrow ((f\;W\;A)\;W)))$. An interesting, non-trivial question
  (suited also for sharpening the intuition on the particular
  conditional logics axioms) is whether these misread statements are
  still provable.  Therefore analogous primed versions have been
  formalized for all correspondence problems as further benchmark
  examples.
 \end{problem}

\begin{problem}\label{Problem5} 
  Do the following logic inclusions hold: (a) CK+\{MP,CS\}
  includes CK+\{CEM\}? (b) CK+\{CEM,MP\} includes CK+\{CS\}? (c) CK+\{RT,AC\} includes CK+\{$(A \Rightarrow B) \rightarrow (((A \wedge B) \Rightarrow C) \leftrightarrow (A \Rightarrow C))$\}? 
The formalizations are obvious and we show only the case for (a):
\begin{align*}
& \forall A,B.  \text{vld}\;\lfloor (A \Rightarrow B) \rightarrow (A \rightarrow B) \rfloor, \\
& \forall A,B. \text{vld}\;\lfloor (A \wedge B) \rightarrow
  (A \Rightarrow B) \rfloor \\
\vdash\quad & \forall A,B. \text{vld}\;\lfloor (A \Rightarrow B) \vee
  (A \Rightarrow \neg B) \rfloor 
\end{align*}
With the results from Problem \eqref{Problem3}, such
questions can alternatively be formalized with the respective semantic
conditions.
\end{problem}

\begin{table}\centering
\begin{tabular}[t]{|l|c|c|c|c|c|c|c|c|c|}
Problem                              & Status & L   & T     & S     & I     & N     & R \\
\hline 
\ref{Problem1}                       & SAT  &       &       & 0.26  &       & 3.34  & 2.88  \\ 
\ref{Problem2}\_RCEA                  & THM  & 0.06  & 0.44  & 0.29  & 18.02 &       &       \\ 
\ref{Problem2}\_RCK                   & THM  & 0.06  & 0.36  & 0.29  & 18.03 &       &       \\ 
\ref{Problem2}\_RCEC                  & THM  & 0.05  & 0.36  & 0.30  & 32.94 &       &       \\ 
\ref{Problem3}\_ID$^\shortleftarrow$    & THM  & 0.03  & 0.36  & 0.27  & 18.03 &       &       \\ 
\ref{Problem3}\_ID$^\shortrightarrow$   & THM  & 0.03  & 0.34  & 0.27  & 17.95 &       &       \\ 
\ref{Problem3}\_MP$^\shortleftarrow$    & THM  & 0.04  & 0.34  & 0.27  & 18.18 &       &       \\ 
\ref{Problem3}\_MP$^\shortrightarrow$   & THM  & 0.04  & 0.35  &       & 18.05 &       &       \\ 
\ref{Problem3}\_CS$^\shortleftarrow$    & THM  & 0.04  & 0.37  & 0.27  & 48.13 &       &       \\ 
\ref{Problem3}\_CS$^\shortrightarrow$   & THM  & 0.13  & 0.37  & 0.27  & 48.42 &       &       \\ 
\ref{Problem3}\_CEM$^\shortleftarrow$   & THM  &       & 1.04  &       & 33.24 &       &       \\ 
\ref{Problem3}\_CEM$^\shortrightarrow$  & THM  & 0.13  & 0.38  &       & 34.10 &       &       \\ 
\ref{Problem3}\_AC$^\shortleftarrow$    & THM  &       & 0.38  & 0.30  & 21.85 &       &       \\ 
\ref{Problem3}\_AC$^\shortrightarrow$   & THM  &       & 0.40  &       & 18.05 &       &       \\ 
\ref{Problem3}\_RT$^\shortleftarrow$    & THM  &       & 0.62  & 1.33  & 18.05 &       &       \\ 
\ref{Problem3}\_RT$^\shortrightarrow$   & THM  &       & 0.42  &       & 18.08 &       &       \\ 
\ref{Problem3}\_CV$^\shortleftarrow$    & THM  &       & 0.44  &       & 125.35&       &       \\ 
\ref{Problem3}\_CV$^\shortrightarrow$   & THM  &       & 0.44  & 86.83 &       &       &       \\ 
\ref{Problem3}\_CA$^\shortleftarrow$    & THM  &       & 0.37  & 1.44  & 18.06 &       &       \\ 
\ref{Problem3}\_CA$^\shortrightarrow$   & THM  &       & 6.83  &       &       &       &       \\ 

\ref{Problem4}\_ID'$^\shortleftarrow$    & THM  & 0.05 & 0.50  & 0.37  & 18.05 &       &       \\ 
\ref{Problem4}\_ID'$^\shortrightarrow$   & THM  & 0.08 & 0.44  & 0.36  & 18.20 &       &       \\ 
\ref{Problem4}\_MP'$^\shortleftarrow$    & THM  & 0.06 & 0.49  & 0.34  & 17.98 &       &       \\ 
\ref{Problem4}\_MP'$^\shortrightarrow$   & CSA  &      &       &       &       & 3.45  & 3.00  \\ 
\ref{Problem4}\_CS'$^\shortleftarrow$    & THM  & 0.04 & 0.36  & 0.29  & 18.01 &       &       \\ 
\ref{Problem4}\_CS'$^\shortrightarrow$   & CSA  &      &       & 0.28  &       & 3.56  & 3.11  \\ 
\ref{Problem4}\_CEM'$^\shortleftarrow$   & THM  &      & 1.04  & 2.00  & 33.23 &       &       \\ 
\ref{Problem4}\_CEM'$^\shortrightarrow$  & CSA  &      &       &       &       & 3.59  & 2.98  \\ 
\ref{Problem4}\_AC'$^\shortleftarrow$    & ?    &      &       &       &       &       &       \\ 
\ref{Problem4}\_AC'$^\shortrightarrow$   & CSA  &      &       &       &       & 4.75  & 3.82  \\ 
\ref{Problem4}\_RT'$^\shortleftarrow$    & THM  &109.42& 31.73 & 0.39  & 18.05 &       &       \\ 
\ref{Problem4}\_RT'$^\shortrightarrow$   & CSA  &      &       &       &       & 3.60  & 3.00  \\ 
\ref{Problem4}\_CV'$^\shortleftarrow$    & THM  &111.06& 31.85 & 0.40  & 51.14 &       &       \\ 
\ref{Problem4}\_CV'$^\shortrightarrow$   & THM  &      & 31.78 & 0.41  & 54.61 &       &       \\ 
\ref{Problem4}\_CA'$^\shortleftarrow$    & CSA  &      &       &       &       & 3.57  & 3.09  \\ 
\ref{Problem4}\_CA'$^\shortrightarrow$   & ?    &      &       &       &       &       &       \\
\ref{Problem5}(a)\_ax.                 & CSA  &      &       &       &       & 3.69  & 3.01  \\
\ref{Problem5}(a)\_sem.                & CSA  &      &       &       &       & 3.62  & 3.06  \\
\ref{Problem5}(b)\_ax.                 & THM  & 0.06 & 8.71  &       &       &       &       \\
\ref{Problem5}(b)\_sem.                & THM  &      &       &       & 63.30 &       &       \\
\ref{Problem5}(c)\_ax.                 & THM  &      & 56.52 &       &       &       &       \\
\ref{Problem5}(c)\_sem.                & THM  &      & 14.89 & 44.69 & 42.46 &       &       \\
\hline
\end{tabular}
\caption{Performance results of HOL-RSs.\label{table1}}
\end{table}

The detailed results of the experiments are presented in
Table~\ref{table1}. Exploiting the System\-On\-TPTP infrastructure,
all experiment runs were done remotely at the University of Miami on
2.80GHz computers with 1GB memory and running the Linux operating
system. The timeout was set to 180 seconds.

The first column of the table presents the problem number and the
second column presents the result status as confirmed by the HOL-RSs:
THM stands for `theorem', CSA for `countersatisfiable', and SAT for
`satisfiable'. The remaining columns report the time after which the
particular HOL-RSs reported the displayed status. L stands for LEO-II,
T for TPS, S for Satallax, I for IsabelleP, N for Nitpick, and R for
Refute.

%
All correspondence claims have been confirmed by the HOL-RSs.
For the primed versions the situation is different and several
counterexamples have been reported by the model finders, in
particular, for several forward directions. 
Two of these counterexamples are exemplary presented next. It is
straightforward to check that they indeed invalidate the respective
primed correspondence statements.\footnote{Concerning, problem
  \ref{Problem4} we might wonder why the suggested denotations for $f$
  below cannot be used as candidates for generating countermodels to
  the corresponding non-primed correspondence statements --- this is
  clearly not the case: e.g., note that the $f$ suggested for
  invalidating MP'$^\shortrightarrow$ (which returns $\emptyset$ for
  all arguments $W$ and $A$) is in fact incompatible with (and thus
  excluded by) MP$^\shortrightarrow$s antecedent (to see this choose
  $A=\{i1\}$, $B=\emptyset$). Such a kind of further analysis is again
  effectively supported by the HOL-RSs.}

Refute reports the following countermodel for MP$^\shortrightarrow$: choose $D_i = \{i1\}$, $A = \{i1\}$, $B =  \{i1\}$, $W = i1$, and 
\begin{small}
 $$f = \left\{
 \begin{array}{ll} 
 i1 & \longrightarrow \left\{
   \begin{array}{ll} 
      \emptyset & \longrightarrow \emptyset \\
      \{i1\}    & \longrightarrow \emptyset \\
   \end{array} \right. \\
 \end{array} 
 \right.
 $$ 
 \end{small}
Nitpick reports for RT'$^\shortrightarrow$: choose $D_i = \{i1,i2\}$, $A = \{i2\}$, $B = \{i1\}$, $C = \emptyset$, $W = i2$, and 
\begin{small}
$$f = \left\{
\begin{array}{ll} 
i1 & \longrightarrow \left\{
  \begin{array}{ll} 
     \emptyset & \longrightarrow \emptyset \\
     \{i1\}    & \longrightarrow \emptyset \\
     \{i2\}    & \longrightarrow \{i2\} \\
     \{i1,i2\} & \longrightarrow \emptyset \\
  \end{array} \right. \\
i2 & \longrightarrow \left\{
  \begin{array}{ll} 
     \emptyset & \longrightarrow \{i2\} \\
     \{i1\}    & \longrightarrow \emptyset   \\
     \{i2\}    & \longrightarrow \{i1\} \\
     \{i1,i2\} & \longrightarrow \emptyset \\
  \end{array} \right. \\
\end{array} 
\right.
$$ 
\end{small}

For both the axiomatic and semantic formalization of Problem
\ref{Problem5}(a) countermodels are quickly found. For example, for the
axiomatic version \ref{Problem5}(a)\_ax. Nitpick reports: choose $D_i
= \{i1,i2\}$, $A = \emptyset$, $B = \{i1\}$, and 
\begin{small}
$$f = \left\{
\begin{array}{ll} 
i1 & \longrightarrow \left\{
  \begin{array}{ll} 
     \emptyset & \longrightarrow \emptyset \\
     \{i1\}    & \longrightarrow \{i1\} \\
     \{i2\}    & \longrightarrow \emptyset \\
     \{i1,i2\} & \longrightarrow \emptyset \\
  \end{array} \right. \\
i2 & \longrightarrow \left\{
  \begin{array}{ll} 
     \emptyset & \longrightarrow \{i1,i2\} \\
     \{i1\}    & \longrightarrow \emptyset   \\
     \{i2\}    & \longrightarrow \{i2\} \\
     \{i1,i2\} & \longrightarrow \{i2\} \\
  \end{array} \right. \\
\end{array} 
\right.
$$ 
\end{small}

 Inclusion claims \ref{Problem5}(b) and \ref{Problem5}(c) are
confirmed as theorems.

With the tools provided by the SystemOnTPTP infrastructure it is
straightforward to write a small shell script which bundles the
mentioned HOL-RSs into a single online reasoning system, and, in fact,
this is how the experiments presented in this section have been
carried out. As the results demonstrate, this combined HOL reasoner is
powerful for reasoning about conditional logics; in particular the
combination of HOL theorem proving and HOL (counter-)model finding is
intriguing. Hence, there is good evidence that the HOL-RSs could
fruitfully support the analysis of similar questions in the
exploration of further conditional logics. Note also, that in most
cases there are at least two matching results by independent
systems. Another interesting observation is that TPS was the strongest
prover in the experiments followed by IsabelleP, Satallax, and
LEO-II. Since this is exactly the opposite order of the outcome of the
2010 CASC\footnote{\scriptsize \url{http://www.tptp.org/CASC/J5/}}
competition, these problems are obviously interesting new benchmarks
for the TPTP library.

The approach is applicable also to reasoning within conditional logics
For example, formula
$ ((p \Rightarrow q) \leftrightarrow (p \rightarrow q)) \rightarrow (p
\Rightarrow p) $ is obviously countersatisfiable, and all model
finders quickly find respective countermodels. Satallax is fastest in
0.28 seconds. The countermodel reported by Nitpick is $D_i=\{i1\},
p=\emptyset, q=\{i1\}$, and
\begin{small}
$$f = \left\{
 \begin{array}{ll} 
 i1 & \longrightarrow \left\{
   \begin{array}{ll} 
      \emptyset & \longrightarrow \{i1\} \\
      \{i1\}    & \longrightarrow \emptyset \\
   \end{array} \right. \\
 \end{array} 
 \right.
$$
\end{small}
Unfortunately, a library of specific benchmark problems for
conditional logics is currently not available, and therefore the (direct)
conditional logics provers CondLean, GoalDuck and leanCK have been
evaluated in \cite{olivetti08:_theor_provin_for_condit_logic} only
with respect to classical modal logic problems. In this
evaluation the modal logic problems were encoded in conditional logics by
defining $\Box \varphi$ as an abbreviation for $T \Rightarrow
\varphi$.
Evaluating our approach wrt. these artificially encoded classical
modal logic problems does hardly make sense though, and the existing
direct embedding of classical modal logic in HOL \cite{J21} should for
good reasons be preferred for these test examples. First
experiments with small hand-translated examples from this test suite
were nevertheless successful.

Evidence against the preconception that our higher-order based
approach to reasoning in conditional logics cannot be effective in
practice comes from a recent case study on automated reasoning in
first-order modal logics
\cite{jens11:_implem_and_evaluat_theor_prover}. In this case study an
higher-order based approach which is closely related to the one
presented here (and which was realized with the provers Satallax and
LEO-II), performed reasonably well behind the specialist provers
MLeanTAB\footnote{\url{http://www.leancop.de/mleantap/programs/mleantap11.pl}}
and
MLeanSeP\footnote{\url{http://www.leancop.de/mleansep/programs/mleansep11.pl}}. In
particular, the higher-order provers did better (in terms of proving
problems) than the direct prover
GQML\footnote{\url{http://cialdea.dia.uniroma3.it/GQML/}} and a
first-order solution based on
MSPASS\footnote{\url{http://www.cs.man.ac.uk/~schmidt/mspass/}}.

As a final remark on this section we underline that solutions to problems in Section 5 are already known in theory of conditional logics. 
This does not straightforwardly imply that HOL-Reasoners (HOL-RSs) can solve them. 
In general, HOL is undecidable, the aim of what presented above is to show that the theoretical embedding of CLs into HOL has also 
practical benefits due that it is possible to use HOL-RSs to reason about and within CLs. This is possible by making HOL-RSs cooperate 
working on the same task. Concerning model generation, we show that countermodels can be constructed for ill-formulated/ill-conjectured correspondence claims.

\section{Conclusion}
A sound and complete embedding of conditional logics into classical
higher-order logic has been presented. Similar to other non-classical
logics \cite{J21}, 
conditional logics can be seen as natural fragments
of classical higher-order logic, and they can be studied and automated as
such. 
Up to authors knowledge, 
the presented work is the first integrated approach to
automated deduction for \emph{all} extensions of CK that involve
\emph{any} combination of the axioms reported in Figure \ref{fig:correspondence}.
Previously existing proof methods \cite{Pozzato10,Pattinson10} are limited to extensions of CK including only  ID, MP, CS and CEM.
Theorem proving for conditional logics appears to be much more difficult than for modal logic. 
There are very few ÒmodalÓ provers for CLs, namely there are sequent calculi for CK+\{ID,MP/CEM,CS\} \cite{Pozzato10} and tableaux for CK+\{CEM,MP\} \cite{Pattinson10}. 
Model builders exist only for CK+\{CEM,MP\} \cite{Pattinson10}. 
No theorem provers are known for  CK+\{CS,AC,RT,CV,CA\} and no model builders are known for CK+\{ID,CS,AC,RT,CV,CA\}. 
The presented methodology offers theorem provers and model finders for above mentioned logics. 
Moreover, the HOL embedding permits use to reason about meta-theorems in an automated way, and naturally extends to First-Order CLs.


Future work includes the systematic analysis of further properties of
conditional logics. For example, following \cite{B12} and motivated by
the results for Problem \ref{Problem5}, the systematic verification (respectively
exploration) of inclusion and equivalence relations between different
conditional logics should be feasible. 
We also plan to create a library of meaningful and challenging
benchmark problems for conditional logics
and 
to evaluate the scalability of our approach. Moreover, a comparison
with direct theorem provers for conditional logics and also with
related techniques based on translations into first-order logic is
needed. However, it is not obvious how these approaches could possibly
be applied for reasoning \textit{about} properties of conditional
logics as studied in this paper. 

Another line of future research is to extend second-order quantifier eliminations techniques like SCAN or SCHEMA to deal with CLs.
In fact, both algorithms are not directly suited for reasoning under CLs due to the peculiarities of selection-function semantics. 

\paragraph{Acknowledgments:} 
We thank the developers and contributors of the TPTP THF
infrastructure and the implementors of the theorem provers and model
finders employed in this work. Christoph Benzmueller is supported by
the German Research Foundation under grants BE 2501/6-1 and BE
2501/8-1.  Valerio Genovese is supported by the National Research
Fund, Luxembourg.

\small
\bibliographystyle{plain} 
\bibliography{Bibliography}
\end{document}